\documentclass{article}

\usepackage{microtype}
\usepackage{graphicx}
\usepackage{subcaption}
\usepackage{booktabs,multirow} % for professional tables

\usepackage{hyperref}

\usepackage[preprint]{icml2026}

\usepackage{amsmath}
\usepackage{amssymb}
\usepackage{mathtools}
\usepackage{amsthm}

\usepackage{enumitem}

\usepackage[capitalize,noabbrev]{cleveref}

\theoremstyle{plain}

\theoremstyle{definition}

\theoremstyle{remark}

\newcommand{\name}{OmniTabBench}

\usepackage{bbding}
\usepackage[most]{tcolorbox}
\tcbuselibrary{breakable, skins, listings}

\newtcblisting{mylisting}{
    listing only,
    breakable,  % Allows page breaks
    enhanced,
    colback=gray!5,
    colframe=gray!50,
    fontupper=\ttfamily\small,
    listing options={
        breaklines=true,
        breakatwhitespace=false,
        basicstyle=\ttfamily\small,
        xleftmargin=10pt,
        xrightmargin=10pt,
        showstringspaces=false
    },
    left=5mm,
    right=5mm,
    top=2mm,
    bottom=2mm,
    arc=2mm
}

\usepackage[textsize=tiny]{todonotes}

\icmltitlerunning{\name{}: Mapping the Empirical Frontiers of GBDTs, NNs, and TFMs for Tabular Data at Scale}

\begin{document}

\twocolumn[
  \icmltitle{\name{}: Mapping the Empirical Frontiers of GBDTs, Neural Networks, and Foundation Models for Tabular Data at Scale}

  % It is OKAY to include author information, even for blind submissions: the
  % style file will automatically remove it for you unless you've provided
  % the [accepted] option to the icml2026 package.

  % List of affiliations: The first argument should be a (short) identifier you
  % will use later to specify author affiliations Academic affiliations
  % should list Department, University, City, Region, Country Industry
  % affiliations should list Company, City, Region, Country

  % You can specify symbols, otherwise they are numbered in order. Ideally, you
  % should not use this facility. Affiliations will be numbered in order of
  % appearance and this is the preferred way.
  \icmlsetsymbol{equal}{*}

  \begin{icmlauthorlist}
    \icmlauthor{Dihong Jiang}{equal,hw}
    \icmlauthor{Ruoqi Cao}{equal,hw}
    \icmlauthor{Zhiyuan Dang}{hw}
    \icmlauthor{Li Huang}{hw}
    \icmlauthor{Qingsong Zhang}{hw}
    \icmlauthor{Zhiyu Wang}{hw}
    \icmlauthor{Shihao Piao}{hw}
    \icmlauthor{Shenggao Zhu}{hw}
    \icmlauthor{Jianlong Chang}{hw}
    \icmlauthor{Zhouchen Lin}{pku}
    \icmlauthor{Qi Tian}{hw}

  \end{icmlauthorlist}

  \icmlaffiliation{hw}{Huawei Cloud}
  \icmlaffiliation{pku}{ School of Intelligence Science
and Technology, Peking University}

  \icmlcorrespondingauthor{Jianlong Chang}{jianlong.chang@huawei.com}

  % You may provide any keywords that you find helpful for describing your
  % paper; these are used to populate the "keywords" metadata in the PDF but
  % will not be shown in the document
%   \icmlkeywords{Machine Learning, ICML}

  \vskip 0.3in
]

% this must go after the closing bracket ] following \twocolumn[ ...

% This command actually creates the footnote in the first column listing the
% affiliations and the copyright notice. The command takes one argument, which
% is text to display at the start of the footnote. The \icmlEqualContribution
% command is standard text for equal contribution. Remove it (just {}) if you
% do not need this facility.

% Use ONE of the following lines. DO NOT remove the command.
% If you have no special notice, KEEP empty braces:
\printAffiliationsAndNotice{}  % no special notice (required even if empty)
% Or, if applicable, use the standard equal contribution text:
% \printAffiliationsAndNotice{\icmlEqualContribution}

\begin{abstract}
  While traditional tree-based ensemble methods have long dominated tabular tasks, deep neural networks and emerging foundation models have challenged this primacy, yet no consensus exists on a universally superior paradigm. Existing benchmarks typically contain fewer than 100 datasets, raising concerns about evaluation sufficiency and potential selection biases. To address these limitations, we introduce \name{}, the largest tabular benchmark to date, comprising 3030 datasets spanning diverse tasks that are comprehensively collected from diverse sources and categorized by industry using large language models. We conduct an unprecedented large-scale empirical evaluation of state-of-the-art models from all model families on \name{}, confirming the absence of a dominant winner. Furthermore, through a decoupled metafeature analysis, which examines individual properties such as dataset size, feature types, feature and target skewness/kurtosis, we elucidate conditions favoring specific model categories, providing clearer, more actionable guidance than prior compound-metric studies.
\end{abstract}

\section{Introduction}
Tabular data are becoming ubiquitous in the real-world, from healthcare \citep{przystalski2023medical, hernandez2022synthetic}, financial \citep{sattarov2023findiff}, meteorological \citep{malinin2021shifts}, to manufacturing industries \citep{zhang2023systematic}, which drives intensive works in understanding and analyzing them. Prior methods for tabular data advance from specialized models including traditional machine learning models (especially tree-based models \citep{chen2016xgboost}) and deep learning models \citep{gorishniy2021revisiting} to general-purpose foundation models \citep{hollmann2025accurate}, where they are developed from a single dataset or across datasets, respectively. While the debate of which models are superior to the others is still ongoing \citep{grinsztajn2022tree, mcelfresh2023neural}, the comparison largely relies on a comprehensive evaluation by running models on a \textit{large} corpus of benchmark datasets. 

Most research papers working in tabular domains seek experimental data from two public repositories, UCI\footnote{https://archive.ics.uci.edu} \citep{KellyUCI} and OpenML\footnote{https://www.openml.org} \citep{vanschoren2014openml}, and one data mining competition platform, i.e. Kaggle\footnote{https://www.kaggle.com}. Existing tabular benchmark datasets are mainly downloaded from one of the three aforementioned sources, yet they only keep fewer than a hundred out of thousands of datasets \citep{erickson2025tabarena, mcelfresh2023neural, grinsztajn2022tree}, which may raise concerns in the sufficiency of evaluation. A prior study in tabular learning \citep{shwartz2022tabular} pointed out that the choice of benchmarking datasets may non-negligibly affect the performance assessment (by introducing biases in the selection of datasets), highlighting the necessity of compiling a large-scale benchmark that is universally suitable for most tabular predictive tasks in various industries, as also emphasized in a recent survey \citep{borisov2022deep}. 

% To better facilitate the evaluation of the tabular machine learning, deep learning, and foundation models in a comprehensive manner, 
To address this limitation, we propose to collect tabular datasets from all three public sources, integrate them into one benchmark which we name \textbf{\name{}}, and categorize the included datasets by industries with the help of large language models (LLMs). To the best of our knowledge, \name{} is by far the largest collection of tabular benchmark datasets (a total of 3030), which is several orders of magnitude (60$\times$ to 375$\times$) greater than most existing benchmarks. 

Furthermore, we revisit the long-standing debate of whether neural networks (NNs) or tree-based models excel in the tabular data modeling by running a few selective models on \textbf{all 3030 datasets}. We confirm that there is still no universal winner that outperforms the other. In addition, we take a closer look at what kind of datasets allow a certain category of models stand out. In contrast to a similar analysis in \citet{mcelfresh2023neural} which explored the correlation between model performance and a compound irregularity metric (that linearly combines a few metafeatures), we choose to decouple the analysis by checking a broader range of metafeatures \textit{individually}, serving as a more clear and practical guidance for researchers to select a proper algorithm when handling different datasets.
Our contributions can be summarized as follows:
\begin{itemize}
    \item We present \name{}, a large-scale and comprehensive benchmark dataset for tabular data, which is greater than existing benchmarks by 60$\times$ to 375$\times$. It contains both classification and regression tasks with varying percentages of categorical and missing values. 
    \item We provide some critical metafeatures of each dataset, such as industry, the ratio of categorical columns, class imbalance, and degree of tailedness, so that users can conveniently access a certain partition of interest by conditioning on corresponding metafeatures.
    \item We evaluate a few state-of-the-art (SOTA) models on \name{} (over 3000 datasets), where each model is from either tree-based models, neural network, or foundation models. It is by far the largest empirical comparison among different models in tabular domain, and we confirm that there is no dominant winner.
    \item We analyze what properties of a dataset may contribute to unleashing the potential of a certain category of models, 
    in order to provide practical insights and references to researchers. Compared to prior related work experimenting with below 200 datasets, our analysis is decoupled, and more reliable with the comprehensive corpus of datasets in \name{}. 
\end{itemize}

\section{Related Work}
\subsection{Debate of NN vs. GBDT in tabular domain} 
% Long before it thrived in modern deep learning, neural network (NN) had been widely applied in machine learning tasks, yet both its performance and interpretablility fell behind in its competition with traditional machine learning counterparts, such as decision tree and support vector machine (SVM). 
Unlike the revolutionary advance in homogeneous data like images \citep{he2016deep} and text \citep{vaswani2017attention}, tabular data pose a significant challenge to deep learning for their heterogeneous nature (e.g. mixed types of variables, different value ranges, missing values) \citep{shwartz2022tabular,arik2021tabnet}, which requires proper preprocessing or transformation to make NN function as expected. Gradient boosted decision tree (GBDT), on the other hand, has been a strong competitor for tabular learning since its invention more than two decades ago \citep{friedman2001greedy}. It utilizes a classic ensemble learning technique, i.e. boosting, to sequentially enhance a weak learner (i.e. decision tree) to a strong ensemble learner. Thanks to many modern implementations of GBDT, such as XGBoost \citep{chen2016xgboost}, CatBoost \citep{prokhorenkova2018catboost}, and LightGBM \citep{ke2017lightgbm}, 
which offer mechanisms that can, for example, deal with categorical/missing values, accelerate the sequential training, and make them scalable to large datasets, 
these optimized GBDTs currently become a desirable option for tabular learning. 
% According to \citet{borisov2022deep}, prior related works approach tabular deep learning from one of three angles: (1) data transformation \citep{sun2019supertml,yoon2020vime}, which encodes tabular input into an NN-friendly format; (2) network design \citep{guo2017deepfm,arik2021tabnet}, which proposes specialized architecture for tabular data; (3) regularization \citep{shavitt2018regularization,kadra2021well}, which enforces regularization to the weights in a network.

% Gradient boosted decision tree (GBDT) has been a strong competitor for tabular learning since its invention more than two decades ago \citep{friedman2001greedy}. It utilizes a classic ensemble learning technique, i.e. boosting, to sequentially enhance a weak learner (i.e. decision tree) to a strong ensemble learner. However, the sequential learning paradigm is inherently inefficient. Thanks to many modern implementations of GBDT, such as XGBoost \citep{chen2016xgboost}, CatBoost \citep{prokhorenkova2018catboost}, and LightGBM \citep{ke2017lightgbm}, which offer mechanisms that can, for example, deal with categorical/missing values, accelerate the sequential training, and make them scalable to large datasets, GBDT currently becomes a desirable (usually the best) option for tabular data modeling. 

The bulk of related studies favor one model over another with different benchmarks. Specifically, \citet{borisov2022deep,shwartz2022tabular,grinsztajn2022tree} found that GBDT achieves better average performance, while \citet{arik2021tabnet,gorishniy2021revisiting,kadra2021well,holzmuller2024better} found NN performs better. It is worth noting that these studies did not use the same tabular data benchmark, and their evaluation datasets were also limited in scale and diversity. For example, \citet{borisov2022deep} used only 5 datasets, and \citet{gorishniy2021revisiting} used 11. \citet{mcelfresh2023neural} benchmarked popular models solely on classification datasets. It is unclear whether and how the conclusion will change if we scale up the number and diversity of evaluation datasets.

In this work, we will revisit this comparison by evaluating a few representative models (including NN and GBDT) on our \name{}.
% newly compiled large-scale (the largest by far) tabular data benchmark, which will be detailed later. 
Note that there is an emerging line of work on tabular foundation models (TFMs) that also use NN, e.g. TabPFN \citep{hollmann2025accurate}. Since its learning paradigm is fundamentally different from the conventional NN training paradigm, we currently categorize it into a separate class of foundation model, but will also include it in our comparison.

\begin{figure*}[!ht]
    \centering
    \includegraphics[width=0.99\linewidth]{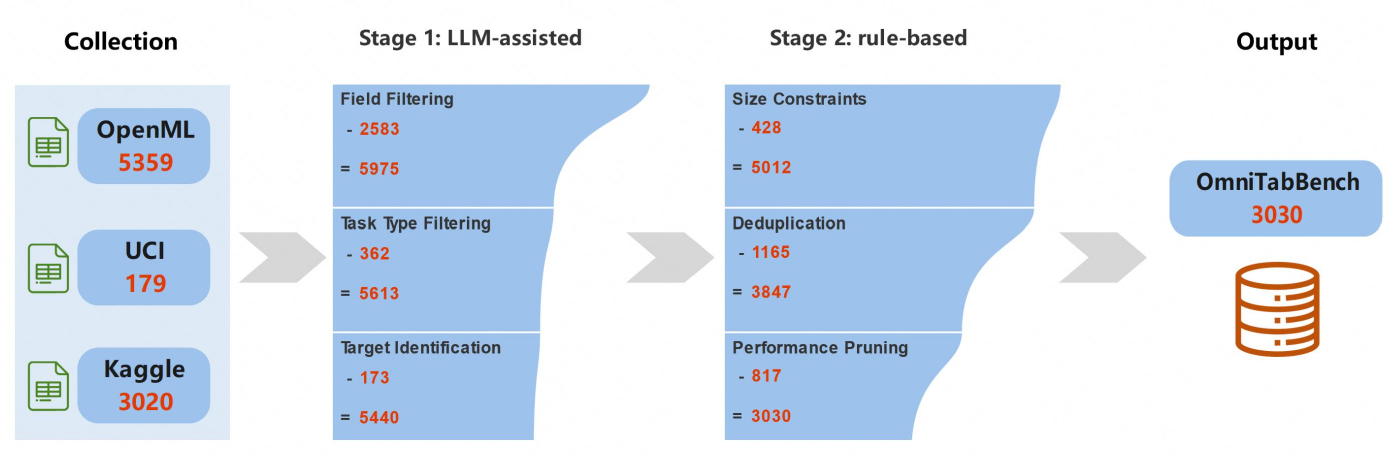}
    \caption{Workflow of constructing \name{}}
    \label{fig:flow_chart}
\end{figure*}

\subsection{Tabular data benchmark} Currently, there is no single source or platform that satisfies the needs of most tabular model development. For example, Kaggle is a machine learning and data mining competition platform, where competition organizers can upload their real-world datasets and problems to seek solutions. However, it is not designed to be a tabular-specific repository, as researchers can upload non-tabular datasets (e.g. images) as well. Besides, raw datasets in both OpenML and Kaggle are not curated, which might be with low utility (e.g. too few rows or columns).

Therefore, building a standardized tabular data benchmark has attracted attention from prior researchers. Existing representative tabular data benchmarks from recent years include TabZilla \citep{mcelfresh2023neural}, TabArena \citep{erickson2025tabarena}, TabReD \citep{rubachev2025tabred}, TableShift \citep{gardner2023benchmarking}, and \citet{grinsztajn2022tree} benchmark.
% , and PMLB \citep{romano2021pmlb}.
Except TableShift and TabRed, the rest of aforementioned benchmarks are in spirit close to our design, which, however, either lack in scale or diversity. For example, TabZilla includes 36 ``hardest" datasets, with classification tasks only. A similar benchmark from \citet{grinsztajn2022tree} includes 45 medium-sized dataset ($\sim$10k). Both of them collect datasets solely from OpenML platform. 
% TabArena is designed to be a living benchmark, where the initial version includes 51 small- to medium-sized (500-250k) datasets. 
% PMLB compiles a total of 450 datasets, but nearly 150 of them contains fewer than 5 columns (where 50 of them only have 2 or even 1 column), which renders its inapplicability in training a machine learning model for restricted feature information.

Despite the existence of these benchmarks, many recent studies in tabular domain still did not choose them in the evaluation. Instead, they evaluate models on their own proposed/collected benchmarks \citep{gorishniy2021revisiting,gorishniy2024tabr, hollmann2025accurate, holzmuller2024better, kim2024carte}. Therefore, in this work, our \name{} aims to bridge this gap by integrating 3030 small- to large-sized datasets 
% (the size of each dataset ranges from 100 to 2 million) 
from diverse sources, which includes both classification and regression tasks in various industries. 

\begin{table*}[!ht]
    \caption{Comparison between \name{} and existing benchmarks. In column ``Sources", ``OML'' represents OpenML, ``New'' means that the authors introduce new datasets from ML products at a company, and ``Public'' means that datasets are collected from other publicly available sources, such as Physionet and American Community Survey. In column ``Tasks'', ``BC'', ``C'', and ``R'' denote binary classification, classification (including multi-class), and regression, respectively. }
    \label{tab:comparsion}
    \centering
    % \scalebox{0.85}{
    \begin{tabular}{lrrrlcc}
    \toprule \multirow{2}{*}{Benchmark} & \multirow{2}{*}{\#Datasets} & \multicolumn{2}{c}{Total Sizes} & \multirow{2}{*}{Sources} & \multirow{2}{*}{Tasks} & \multirow{2}{*}{Categorize?} \\
    \cmidrule(lr){3-4} 
    
         &  & \#Rows & \#Cols  &  \\
    \midrule
    %%% median sizes
    % TabZilla \citep{mcelfresh2023neural} & 36 & 3087 & 23 & OML & C & \XSolidBrush\\
    % TabArena \citep{erickson2025tabarena} & 51 & 6497 & 20 & OML, UCI, Kaggle & C, R & \XSolidBrush\\
    % \citet{grinsztajn2022tree} benchmark & 45 & 16679 & 13 & OML & C, R &  \XSolidBrush\\
    % PMLB \citep{romano2021pmlb} & 450 & 1000 & 25 & OML, UCI, Kaggle & C, R& \XSolidBrush\\
    % TabReD \citep{rubachev2025tabred} & 8 & 7163150 & 261 & Kaggle, Own & C, R & \XSolidBrush\\
    % TableShift \citep{gardner2023benchmarking} & 15 & 840582 & 23 & Public & BC & \XSolidBrush\\
    
    %%% total sizes
    % PMLB \citep{romano2021pmlb} & 450 & 18.0M & 8.7k & OML, UCI & C, R& \XSolidBrush\\
    TableShift \citep{gardner2023benchmarking} & 15 & 17.0M & 15.6k & UCI, Kaggle, Public & BC & \XSolidBrush\\
    TabArena \citep{erickson2025tabarena} & 51 & 1.0M & 6.0k & OML, UCI, Kaggle & C, R & \XSolidBrush\\
    \citet{grinsztajn2022tree} benchmark & 45 & 10.9M & 2.1k & OML & C, R &  \XSolidBrush\\
    TabReD \citep{rubachev2025tabred} & 8 & 2.2M & 3.0k & Kaggle, New & C, R & \XSolidBrush\\
    TabZilla \citep{mcelfresh2023neural} & 36 & 2.4M & 8.2k & OML & C & \XSolidBrush\\
    \midrule
    \name{} & \bf 3030 & \bf 144M & \bf 666k & OML, UCI, Kaggle & C, R & \Checkmark \\
    \bottomrule
    \end{tabular}
    % }
\end{table*}

\section{Construction of \name{}}
% \Cref{fig:flow_chart} illustrates the overall workflow of our benchmark construction and filtering results per step.
\subsection{Dataset collection}
Our main data sources are from OpenML, UCI and Kaggle. Both OpenML and UCI repositories are designed for machine learning research and primarily stores tabular datasets, so we collect all available datasets by looping over all valid indices (each dataset has a unique ``idx"). For Kaggle, we query it with ``machine learning" as the keyword and collect all the returned datasets. From these three public sources combined, we initially collect a total of 8558 datasets (5359 from OpenML, 3020 from Kaggle, and 179 from UCI). Since raw datasets may be duplicated and may not be desired for practical use, a filtering procedure is mandatory before integrating them into our benchmark. Given the amount of collected datasets, the manual screening will be labor-intensive and inefficient. Therefore, we set a few pre-defined rules to refine and categorize our initial collection of raw datasets, with the help of an LLM.

\subsection{LLM-assisted screening and categorization}
\label{sec:llm-assist}
Meta-information of each dataset, e.g. type of data or the industry that data come from, is not always clearly tagged or presented. Instead, the meta-information is generally hidden in the long text description of each dataset, or implied by the column names. Thanks to the advance of modern LLMs, we propose to delegate the initial screening and categorization of all datasets to an LLM via detailed instructions/prompts. 

Specifically, we ask an LLM, i.e. doubao (doubao-1-5-pro-32k-character-250715), to extract key information from two text sections, i.e. the ``About Dataset'' section and the file or column description (if there is), and encode the structured information in a separate file for future reference, including:

\textbf{Field:} indicating which field the dataset falls into. Label ``CV", ``NLP", or ``ML" if it can be clearly determined, otherwise label ``OTHERS". We mainly keep ``ML'' datasets (5975 datasets remain). 

\textbf{Task type:} indicating which kind of task this dataset focuses on. Choose ``classification'' or ``regression'' if the task type can be clearly determined, otherwise label ``uncertain''. We only keep classification and regression tasks (5613 datasets remain).

\textbf{Target column:} identifying target column name(s) from the given column names whenever explicitly mentioned, otherwise label ``uncertain". Datasets with uncertain target column names will be removed from our collection (5440 datasets remain).

\textbf{Industry:} indicating which industry the data come from. Choose one of ``Internet and Web", ``Physical Science", ``Business and Finance", ``Health and Fitness", ``Earth and Environment", ``People and Society", ``Energy and Industry", ``Arts and Media", and ``Others". We mainly use this information to categorize all datasets into their domains.

We give an example of full prompts in \Cref{sec:full_prompt}.
% Limitation: the LLM screening may not be perfect, for example, LLM specifies the incorrect domain or target column. Currently we choose to trust the results returned by LLM, for failing to discover a refining method without manual endeavor, which could be a viable angle for future work.

% another: subsampled datasets (\#rows and \#cols are subsampled from a large dataset) are not de-duplicated.

\subsection{Rule-based filtering}
\label{sec:filtering}
After the initial screening, we further filter the rest datasets based on the following factors:

\textbf{Size:} To remove datasets with limited information, we only keep the datasets with the number of features/columns between 5 and 2000 and the number of observations/rows between 100 and 2 million (5012 datasets remain).

\textbf{Deduplication:} Since we collect datasets from various sources, there might be some datasets that are repetitively submitted. We design a fingerprint for each dataset as an indicator for deduplication from the following three steps:
\begin{itemize}[leftmargin=*]
    \item \textbf{Column names:} Duplicate datasets have the same column names. To account for the reformatting or reorganization of columns, we remove all non-alphabetic and non-digit characters in column names, sort them based on the alphabetic order, and concatenate them into one string. This string should be identical for duplicate datasets.
    \item \textbf{Number of rows and columns:} The step above does not necessarily remove all duplicate datasets, especially when the number of columns is small, so we consider the number of rows and columns as an additional factor. We convert the number of rows and columns of a dataset into a string.
    \item \textbf{Target columns:} In rare cases, however, the above two steps still cannot assure the removal of all duplicated datasets. Therefore, we resort to the distribution of the target column as the final component of the fingerprint. Specifically, for continuous target columns, we compute deciles (i.e. 0.1-, 0.2-, ..., 0.9-quantile) combined with min and max values; for discrete target columns, we count the number of instances in each class and sort them. The numerical values are converted into strings and joined together.
\end{itemize}
Finally, we concatenate the three strings from the above steps into one, and get its hash value as a digital fingerprint. Identical hash values imply duplication of datasets (3847 datasets remain). Examples of our designed fingerprints can be found in \Cref{sec:fingerprint}.

\textbf{Difficulty:} We quantify the difficulty of a predictive task based on the performance of four selected models (see \Cref{sec:models}). We use F$_1$ score and R$^2$ (coefficient of determination) to evaluate the performance of selected models on classification and regression tasks, respectively. Our intention is to remove datasets that are either too easy or too hard (or mistakenly chosen target columns by LLM), thus we only keep datasets that any one of four selected benchmarked models achieves an F$_1$ score or R$^2$ between 0.1 and 0.9 (3030 datasets remain).
% F1 too high -> too easy; F1 too low -> too hard / something wrong with the dataset
    
% It is worth noting that we circumvent the \textbf{license} issue by not publishing a new dataset. Instead, we provide the aggregated information, e.g. web/download link, industry, metafeatures, of all datasets. Users only need to download the dataset from the link we record.

\begin{figure*}[t]
\centering
    \begin{subfigure}{0.50\linewidth}
        \includegraphics[width=\linewidth,keepaspectratio=true]{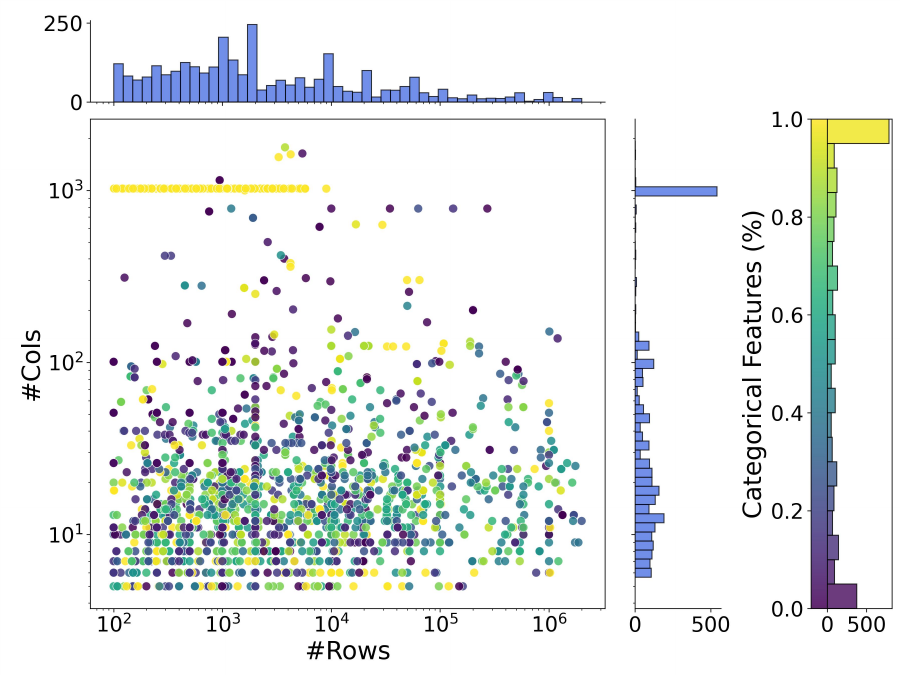}
        \caption{Overview of \name{}}
        \label{fig:omnitabbench}
    \end{subfigure}
    \hspace{2ex}
    \begin{subfigure}{0.43\linewidth}
        \includegraphics[width=\linewidth,keepaspectratio=true]{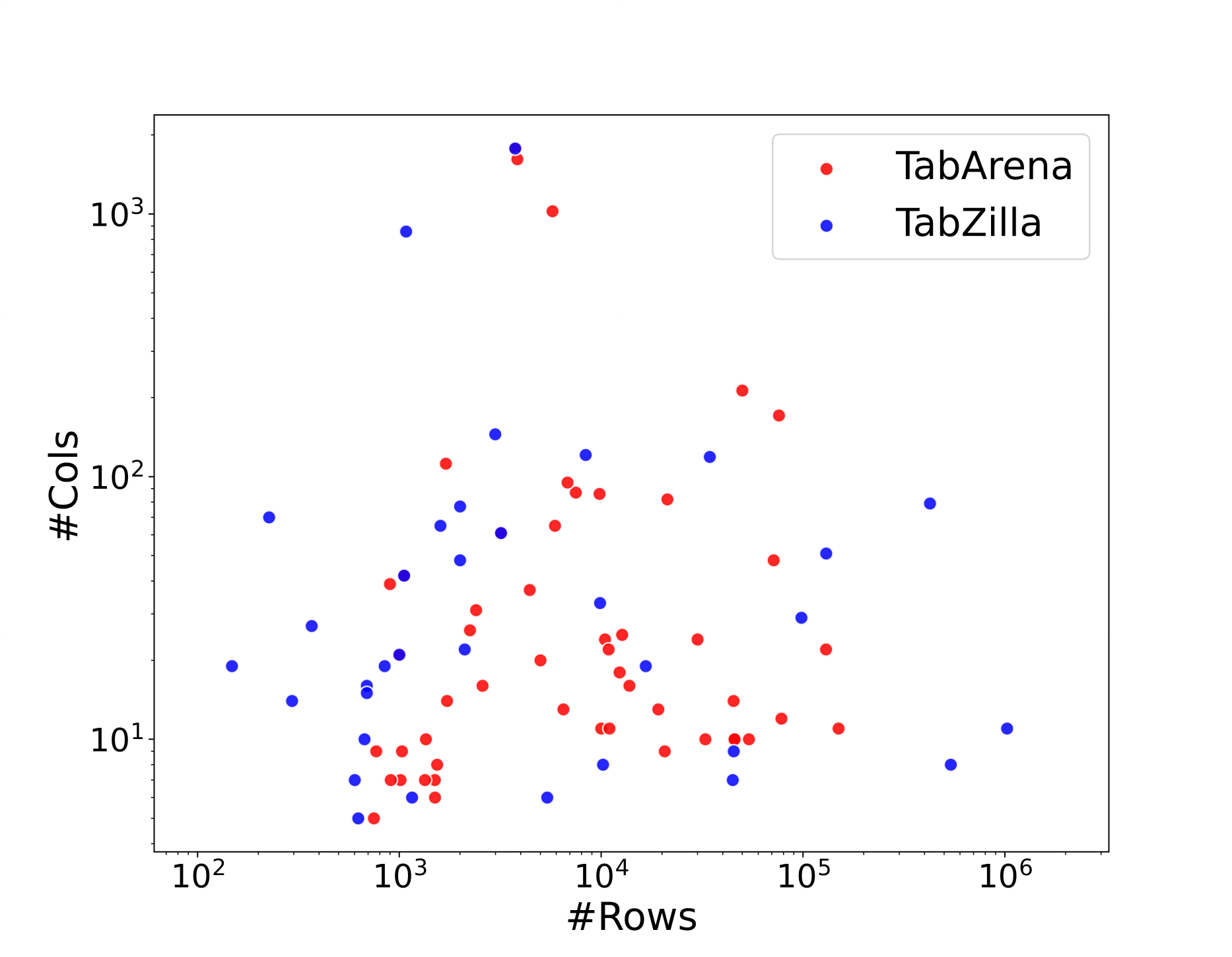}
        \caption{Overview of two existing benchmarks}
        \label{fig:other_benchmark}
    \end{subfigure}
\caption{Comparison between \name{} and existing representative benchmarks. (a) We visualize the number of rows, columns, and the percentage of categorical columns per dataset in \name{}, as well as their distributions. (b) TabZilla and TabArena contain notably fewer datasets than \name{}.}
\label{fig:complex_scatter}
\end{figure*}

\begin{figure}[t]
    \centering
    \includegraphics[width=0.98\linewidth]{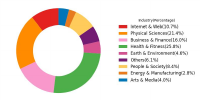}
    \caption{Categorization of \name{} by industries}
    \label{fig:industry}
\end{figure}

\subsection{Comparison with existing benchmarks}
We summarize the critical dimensions of comparison between existing benchmarks and \name{} in \Cref{tab:comparsion}. Notably, the scale of \name{}, in terms of the number of datasets, the number of rows and columns, exceeds existing benchmarks by a sizable margin. For example, though TabArena is also one of the few datasets that seek data from diverse sources and contain diverse tasks, the scale of \name{} is significantly larger than TabArena, as the total number of datasets, rows, and columns exceed TabArena by 60$\times$, 144$\times$, and 111$\times$, respectively. \Cref{fig:complex_scatter} illustrates an overview of \name{} and comparison with two existing benchmarks.

Besides, researchers from a specific industry may want to use experimental data from that particular domain, as the data pattern or distributions may vary from industry to industry. However, most existing benchmark datasets fail to provide such information. In contrast, our \name{} offers a coarse-grained categorization by industries for all 3030 datasets, which spans a wide range of industries as detailed in \Cref{fig:industry}.

\begin{figure*}[!ht]
    \centering
    \includegraphics[width=0.78\linewidth]{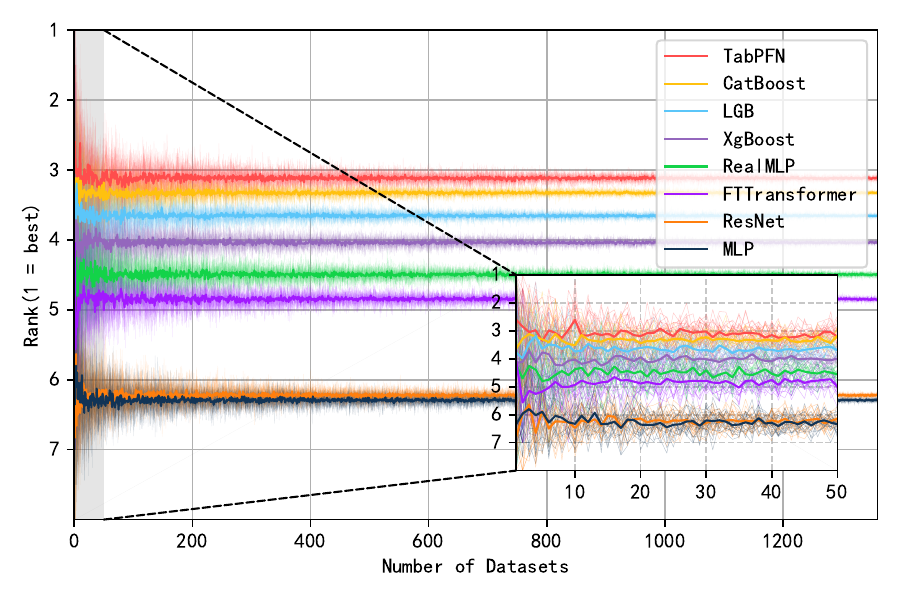}
    \caption{Rank of different models with increasing the number of evaluation datasets. The zoom-in window takes a closer look at the rank variation with a limited number of datasets. We plot this figure from datasets that all eight models have results.}
    \label{fig:rank_across_dataset}
\end{figure*}

\section{Experiments}
\subsection{Experimental Setup and Selected Models}
\label{sec:models}
\textbf{Benchmarked Models: } In light of the scale of our collected datasets, it is too costly to run experiments with many benchmarked models in an exhaustive manner as is completed in \citet{mcelfresh2023neural,erickson2025tabarena}. Instead, we choose a few representative models in our experiments: (1) GBDTs: LightGBM (LGB) \citep{ke2017lightgbm}, XGBoost \citep{chen2016xgboost} and CatBoost \citep{prokhorenkova2018catboost}; (2) NNs: a classicial two-layer MLP, RealMLP \citep{holzmuller2024better}, ResNet \citep{gorishniy2021revisiting}, and FT-Transformer \citep{gorishniy2021revisiting};
% a recent carefully tuned MLP that optimizes the vanilla MLP with a series of engineering tricks, which makes it on a par with GBDT in performance; 
(3) TabPFN \citep{hollmann2025accurate}, a transformer-based large tabular model pretrained on millions of synthetic datasets, which initiates and stimulates research in TFMs. We provide more implementation details in \Cref{sec:implementation}.

We choose these models because they indicate SOTA or near SOTA performance in tabular learning as reported in many related literatures \citep{gorishniy2021revisiting, hollmann2025accurate, rubachev2025tabred}, and each of them belongs to a representative learning paradigm. Nevertheless, we welcome researchers and engineers to evaluate their developed methods on our benchmark, regardless of learning paradigm. Note that we aim to test these models' \textbf{out-of-the-box} capability, therefore we use default or standard configurations of these models without tuning them.

Taking model performance, learning paradigm, and training efficiency all into account, we select LGB, MLP, RealMLP, and TabPFN to quantify the difficulty of all filtered datasets (see Difficulty in \Cref{sec:filtering}).

\textbf{Predictive tasks: } We consider classical tabular predictive tasks, i.e. classification (including both binary and multi-class classification) and regression. We use the same metrics that quantify the difficulty of each dataset as described in \Cref{sec:filtering} to evaluate the performance of selected models.

\textbf{Experimental setup: } We train and evaluate eight benchmarked models on all filtered datasets to indicate that our benchmark could serve as a comprehensive repository for developing and comparing tabular machine learning models. Besides, with the development of emerging synthetic-data-pretrained tabular foundation models, our benchmark also satisfies the needs when one wants to evaluate such models like TabPFN on real-world datasets. For example, while \citet{hollmann2025accurate} evaluates and compares TabPFN across around 140 real-world datasets, we extend this evaluation by running their publicly released pretrained model on our benchmark. It is worth noting that TabPFN is designed to handle small- to medium-sized datasets with up to 10k samples and 500 features. Therefore, we evaluate TabPFN on a subset (1815 datasets) of our benchmark, which is already 12$\times$ more than their evaluated real datasets.
 %(TabPFN only evaluates on 28+29+36+45+5=143 datasets)

\subsection{Preprocessing}
% for all models? 
While TabPFN and many GBDT implementations include various built-in preprocessors for different variables and missing values, the vanilla MLP is not inherently equipped with similar techniques, and the lack of those preprocessing techniques may lead to drastic performance drop for neural network like MLP. Therefore, we perform a few preprocessing steps to the raw dataset to make the empirical comparison fair for all models. We detail the preprocessing for different values below.

\textbf{Missing values:} We keep this step the same for all models. Infinity and missing values in numerical columns are imputed by the average values in corresponding columns. Missing values in categorical columns are temporarily imputed by a placeholder, which is treated as a special categorical variable.

\textbf{Numerical values:} We keep this step the same for all models. (1) We transform low-cardinality numerical columns into categorical columns. We define low-cardinality columns as those columns with the number of unique values (\#unique) below 20 and the ratio of \#unique over \#total\_rows below 0.2. (2) For the rest numerical columns, we clip each column to a range between its 0.05-quantile and 0.95-quantile, then apply quantile transform (QuantileTransformer in Scikit-learn \citep{pedregosa2011scikit}) to them. 

\textbf{Categorical values:} For all benchmarked models except vanilla MLP, we simply apply label encoding (that maps discrete variables to natural numbers, which can be completed by LabelEncoder in Scikit-learn) to all categorical columns, since they are equipped with categorical variable processing mechanisms. For MLP, we apply one-hot encoding to categorical columns with cardinality below 50 (\#unique $\leq$ 50) and label encoding to the rest categorical columns.

% low-cardinality numeric -> categorical

% categorical impute, labelEncoding

% numeric, infinity and missing value -> both impute as average, remove uppper and lower 5\%-> quantiletransform

% categorical -> nunique <= 50: onehotencoding, nunique > 50: labelencoding (only for mlp estimator)

\begin{table*}[!ht]
    \caption{The mean metafeatures of datasets that each category of models excel. We use $\gamma, \kappa$ to denote skewness and kurtosis, respectively. We compute p-values via Welch's t-test in the bottom three rows, where the statistically significant ($p<0.05$) pairwise difference is highlighted in bold (p-values lower than 1e-3 are rounded to 0).}
    \label{tab:metafeatures}
    \centering
    % \scalebox{0.85}{
    \begin{tabular}{lcccccccccc}
    \toprule \multirow{2}{*}{Rank-1 Models} & \multicolumn{3}{c}{Sizes} & \multicolumn{2}{c}{\small Feature Types} & \multicolumn{2}{c}{\small Feature Distribution} & \multicolumn{3}{c}{\small Target Distribution} \\
    \cmidrule(lr){2-4} \cmidrule(lr){5-6} \cmidrule(lr){7-8} \cmidrule(lr){9-11} 
    
         & $\#R$ & $\#C$ & $\#R/\#C$ & $\%Cat.$ & $\%Mis.$ & $|\gamma|$ & $\kappa$ & Entropy & $|\gamma|$ & $\kappa$ \\
    \midrule
    GBDT & 6143.5 & 40.3 & 349.7 & 0.437 & 0.011 & 0.796 & 1.510 & 0.800 & 0.780 & 0.431\\
    NN & 8259.7 & 39.1 & 399.4 & 0.574 & 0.016 & 0.645 & 1.241 & 0.754 & 0.951 & 1.433\\
    TabPFN & 2773.2 & 36.0 & 166.2 & 0.368 & 0.014 & 0.625 & 0.524 & 0.860 & 0.687 & 0.126\\
    \midrule
    GBDT vs NN & \bf 0 & 0.841 & 0.265 & \bf 0 & 0.052 & \bf 0 & 0.764 & \bf 0.009 & 0.108 & 0.083\\
    TabPFN vs NN & \bf 0 & 0.527 & \bf 0 & \bf 0 & 0.442 & 0.644 & 0.343 & \bf 0 & \bf 0.007 & \bf 0.019\\
    GBDT vs TabPFN & \bf 0 & 0.299 & \bf 0 & \bf 0 & 0.227 & \bf 0 & 0.091 & \bf 0 & 0.185 & 0.210 \\
    \bottomrule
    \end{tabular}
    % }
\end{table*}

\subsection{Results}
\subsubsection{Large-scale benchmark is necessary}
Inspired by Figure 1 in \citet{kohli2024towards}, we visualize the impact of the number of evaluation datasets in \Cref{fig:rank_across_dataset}. The semi-transparent thin lines correspond to the rank of different models on randomly sampled subsets of \name{}. We take 20 random subsets at each number, and the solid line corresponds to the average rank over random subsamples. Although the rank (or relative order) of different models converges as we scale up the evaluation datasets, it is worth noting that the rank over random subsets drastically oscillate when the number of evaluation datasets is limited. As highlighted in the zoom-in window in \Cref{fig:rank_across_dataset}, some NNs or GBDTs may occasionally surpass all other models with under 50 datasets, despite TabPFN ranks best on average, which suggests that insufficient evaluation may introduce bias in the selection of datasets and therefore lead to misleading conclusions.

We add a couple of remarks based on our results: (1) Besides competitive performance, GBDTs are significantly faster than NNs, thus we suggest GBDT can always be a fast sanity check protocol before developing any complex models; (2) Vanilla MLP is not a competitive baseline, but its well-tuned counterpart, i.e. RealMLP, significantly improves the performance of MLP, showcasing the potential of NNs.

\subsubsection{Still no dominant winner}
To compare the potential ceiling capability of different category of models, we aggregate results by \textbf{taking the max} performance score per category, and compare them on the subset of 1815 datasets that the pretrained TabPFN can infer on. A bit surprisingly, GBDT, NN, and TabPFN wins (ranks the best) on 31.6\%, 33.9\%, and 34.5\% of the 1815 datasets, respectively.
% the proportion of datasets that each category of models ranks the first in \Cref{tab:best_by_model}. 
Although \Cref{fig:rank_across_dataset} illustrates a clear trend that TabPFN $>$ GBDT $>$ NN on average, the ensemble of NNs achieves a three-way near-tie with GBDTs and TabPFN, which indicates that there is still no dominant winner among these models. This result also motivates us to further examine the intrinsic patterns in the datasets that each category of models excel. 

% \begin{table}[t]
%   \caption{The proportion of datasets that each category of models achieve the best performance. We here adopt the subset of 1815 datasets that the pretrained TabPFN can infer on.}
%   \label{tab:best_by_model}
%   \begin{center}
%     % \begin{small}
%     %   \begin{sc}
%         \begin{tabular}{lccc}
%           \toprule
%           Models  & GBDT & NN & TabPFN    \\
%           \midrule
%           Win rate & 31.6\% & 33.9\% & 34.5\%  \\
%           \bottomrule
%         \end{tabular}
%     %   \end{sc}
%     % \end{small}
%   \end{center}
%   \vskip -0.1in
% \end{table}

% \begin{figure}[t]
%     \centering
%     \includegraphics[width=0.98\linewidth]{figs/rank_best_models_chart.pdf}
%     \caption{The proportion of datasets that each category of models achieve the best performance}
%     \label{fig:best_by_model}
% \end{figure}

\begin{figure*}[!ht]
    \centering
    \includegraphics[width=0.98\linewidth]{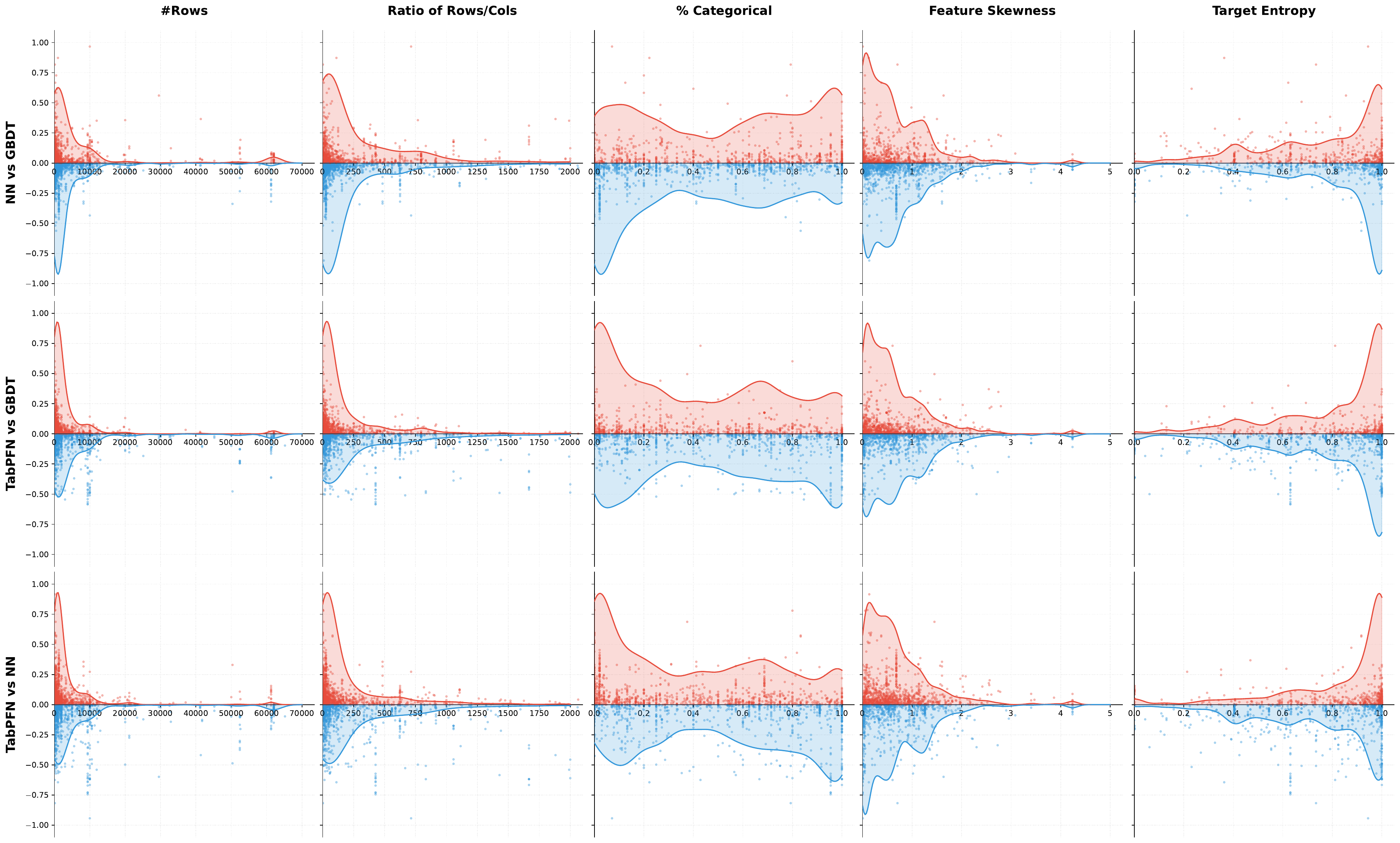}
    \caption{The distribution of performance gap between different pairs of models. Five columns represent five different metafeatures, and three rows denote three pairwise comparison. Performance gap on different datasets refers to the subtraction of the score between the former and latter models (for example, NN vs GBDT means subtracting score of GBDT from NN), which are quantified by red (positive/win) and blue (negative/loss) points, respectively. We also fit a PDF of the points along each varying metafeatures via kernel density estimation.}
    \label{fig:performance_gap_comp}
    \vspace{-2ex}
\end{figure*}

\subsubsection{Metafeature analysis}
Intuitively, datasets with various characteristics may benefit the learning of different models as analyzed in \citet{mcelfresh2023neural}, where the authors compute the correlation between the performance gap (of NNs vs GBDTs) and a linear combination of a few metafeatures of datasets. Differently, we decouple the analysis by checking how the individual metafeature is associated with the success of each model, which is more straightforward and indicative. We summarize the considered metafeatures as follows:
\begin{itemize}[leftmargin=*]
    \item \textbf{Size:} the number of rows and columns, and their ratio.
    \item \textbf{Feature types:} the ratio of categorical and missing values.
    \item \textbf{Feature distribution:} we compute the third and fourth moments of feature columns, i.e. skewness and kurtosis. They are widely accepted to measure the degree of asymmetry and tailedness of a distribution, respectively. We compute absolute-valued skewness because we care how skew the distribution is instead of its direction.
    \item \textbf{Target distribution:} as an indicator for imbalance, we compute Shannon entropy for categorical target columns, and skewness/kurtosis for continuous target columns.
\end{itemize}

% \paragraph{Size:} the number of rows and columns, and their ratio.
% \paragraph{Feature types:} the ratio of categorical features and missing values.
% \paragraph{Feature distribution:} we compute the third and fourth moments of feature columns, i.e. skewness and kurtosis. They are widely accepted to measure the degree of asymmetry and tailedness of a distribution, respectively. We compute absolute-valued skewness because we care how skew the distribution is more than whether it is left or right skewed.
% \paragraph{Target distribution:} as an indicator for imbalance, we compute Shannon entropy for categorical target columns, and skewness/kurtosis for continuous target columns.

\cref{tab:metafeatures} summarizes the centroid of metafeatures of datasets where different models ranked number one, which reveals that the optimal model choice can be governed by the structural characteristics of the dataset. NNs demonstrate a distinct advantage in "high-information" regimes, characterized by larger sample sizes ($\#R=8259.7$), a higher data density ($\#R/\#C \approx 400$), and a higher proportion of categorical features ($\%Cat.=0.574$), all with $p < 0.001$. This result suggests that by mapping categorical features into a differentiable, low-dimensional embedding space, unlike the sparse one-hot or statistical encodings common in tree-based models, modern embedding techniques enable NNs to learn flexible, smooth representations of complex interactions, particularly when sufficient data is available.
% Furthermore, the NN group displays a tolerance for higher target skewness ($|\gamma|=0.951$) and kurtosis ($\kappa=1.433$), indicating that deep architectures are statistically more robust to extreme target distributions when sufficient data density is available to facilitate convergence.

Conversely, GBDTs maintain a critical niche in handling dataset "irregularity." Our analysis shows that GBDTs are the superior choice for datasets with significantly higher feature skewness ($|\gamma|=0.796$) and kurtosis ($\kappa=1.510$) compared to those where NNs or TabPFN excel. The statistical significance of feature skewness of datasets where GBDT or NN/TabPFN wins ($p < 0.001$) underscores a fundamental advantage of the tree-based inductive bias: because trees are rank-based and rely on ordinal splits, they are naturally more robust to skewed or heavy-tailed feature distributions, which makes GBDTs the more reliable baseline for raw or unnormalized tabular data often encountered in the real world. This result also resonates with prior smaller-scaled study that GBDT fits irregular functions better than NNs \citep{grinsztajn2022tree}.

Finally, TabPFN achieves the SOTA performance for the low-sample frontier, winning on datasets with the lowest average row count ($\#R=2773.2$) and the lowest row-to-column ratio ($\#R/\#C \approx 166$), because it is designed for smaller datasets (restricted by the context length of transformers). This result combined with the best average rank of TabPFN in \cref{fig:rank_across_dataset} suggests that TabPFN is likely the best option when the dataset fits its capability.
In addition, TabPFN's dominance is concentrated in datasets with the lowest target skewness ($|\gamma|=0.524$) and kurtosis ($\kappa=0.126$), suggesting that while its in-context prior is highly effective for rapid generalization on small data, its performance is most stable when the target distribution is relatively regular, which is also consistent with the guide from the original paper \citep{hollmann2025accurate} that TabPFN favors smooth regression datasets.

To give a more detailed overview of model comparison with varying metafeatures, we also visualize in \Cref{fig:performance_gap_comp} the distribution of performance gap between different pairs of models with five selected metafeatures (with at least two out of three statistically significant pairwise difference). The trends and patterns are consistent with results in \Cref{tab:metafeatures}. For example, TabPFN showcases its superiority over NN and GBDT on smaller-scaled datasets (fewer rows and smaller row-to-column ratio), and NN's advantage against GBDT and TabPFN shows a positive correlation with increasing categorical ratios. The full pairwise comparison on all metafeatures are deferred to \Cref{sec:more_gaps}.
% \red{what conclusions can we draw?}
% \red{features with no statistically significant pvalues worth noting too. They are not informative metafeatures?}

\section{Conclusion}
In this work, we have presented \name{}, a groundbreaking large-scale benchmark that aggregates 3030 tabular datasets from major public sources, surpassing existing collections by orders of magnitude and incorporating industry-specific categorizations facilitated by LLMs. Through exhaustive experiments on this comprehensive corpus, we reaffirm that no single modeling paradigm (GBDTs, NNs, or foundation models) universally outperforms the others in tabular predictive tasks, which underscores the persistent "no free lunch" theorem and highlights the context-dependent nature of model efficacy.

Our decoupled metafeature analysis provides actionable insights into dataset properties that correlate with superior performance for each model category, including sizes, categorical and missing rate, feature and target distributional characteristics (e.g., skewness and kurtosis).
% For instance, GBDTs exhibit robustness on moderately sized datasets with skewed features, neural networks leverage higher categorical proportions and larger scales, and foundation models like TabPFN excel in small, symmetric regimes. 
Our observations, derived from a far broader empirical foundation than prior studies, offer practical guidance for algorithm selection in real-world applications.
By releasing \name{} with detailed metafeatures and evaluation results, we aim to mitigate benchmarking biases, foster reproducible research, and stimulate further innovations in tabular modeling. 
% Future directions include expanding to multimodal tabular data, incorporating dynamic or streaming tasks, and developing meta-learning frameworks that automatically recommend models based on metafeatures. Ultimately, this benchmark paves the way for more reliable progress toward versatile and industry-aligned tabular intelligence.

\bibliography{references}
\bibliographystyle{icml2026}

%%%%%%%%%%%%%%%%%%%%%%%%%%%%%%%%%%%%%%%%%%%%%%%%%%%%%%%%%%%%%%%%%%%%%%%%%%%%%%%
%%%%%%%%%%%%%%%%%%%%%%%%%%%%%%%%%%%%%%%%%%%%%%%%%%%%%%%%%%%%%%%%%%%%%%%%%%%%%%%
% APPENDIX
%%%%%%%%%%%%%%%%%%%%%%%%%%%%%%%%%%%%%%%%%%%%%%%%%%%%%%%%%%%%%%%%%%%%%%%%%%%%%%%
%%%%%%%%%%%%%%%%%%%%%%%%%%%%%%%%%%%%%%%%%%%%%%%%%%%%%%%%%%%%%%%%%%%%%%%%%%%%%%%
\newpage
\appendix
\onecolumn
% \section{You \emph{can} have an appendix here.}

% You can have as much text here as you want. The main body must be at most $8$
% pages long. For the final version, one more page can be added. If you want, you
% can use an appendix like this one.

% The $\mathtt{\backslash onecolumn}$ command above can be kept in place if you
% prefer a one-column appendix, or can be removed if you prefer a two-column
% appendix.  Apart from this possible change, the style (font size, spacing,
% margins, page numbering, etc.) should be kept the same as the main body.
%%%%%%%%%%%%%%%%%%%%%%%%%%%%%%%%%%%%%%%%%%%%%%%%%%%%%%%%%%%%%%%%%%%%%%%%%%%%%%%
%%%%%%%%%%%%%%%%%%%%%%%%%%%%%%%%%%%%%%%%%%%%%%%%%%%%%%%%%%%%%%%%%%%%%%%%%%%%%%%
\section{Implementation}
\label{sec:implementation}
\subsection{Model Details}
We adhere to the principle of prioritizing publicly available or widely recognized versions from official repositories for implementing our selected models, while resorting to custom implementations only when no standard version is available to meet our specific architectural requirements.

For three GBDT implementations, including XGBoost, LightGBM and CatBoost, we utilize their official Python packages, employing default hyperparameter configurations (with default number of estimators = 100) to ensure reproducibility.

% The LightGBM model is implemented through the official lightgbm Python package, following the standard framework for gradient boosting decision trees.

% Regarding CatBoost, we employ the implementation provided by the catboost Python package, specifically leveraging its native support for categorical features to optimize performance on structured data.

In terms of ResNet and FT-Transformer, we adopt the implementations provided by \citet{gorishniy2021revisiting} from the official repository (\url{https://github.com/yandex-research/rtdl-revisiting-models}), as these architectures represent the state-of-the-art in deep learning for tabular and structured data.

For the RealMLP model, we utilize the architecture as implemented in the pytabkit framework \citep{holzmuller2024better} to serve as a refined neural baseline for our comparative analysis. We adopt the official model configurations and hyperparameters.

As for our Hierarchical MLP, we employ a specifically structured multilayer perceptron (two hidden layers of 256 neurons) designed for hierarchical feature transformation where input embeddings are mapped through a sequence of linear projections and Rectified Linear Unit (ReLU) activations, followed by a 50\% dropout layer ($p=0.5$) for regularization and a dedicated linear head for task-specific prediction.

We employ the official implementation of TabPFN from the PriorLabs repository (\url{https://github.com/PriorLabs/TabPFN}), specifically utilizing the tabpfn-v2-regressor.ckpt and tabpfn-v2-classifier.ckpt versions for our regression and classification tasks, respectively.

\subsection{Data Partitioning}

To ensure a robust evaluation across diverse datasets, we implement an automated data partitioning pipeline that dynamically adapts its splitting logic based on the identified task type of each target column.

In cases where the dataset contains \textbf{multiple target variables}, the system iterates through each column to treat them as entirely independent tasks, ensuring that the feature-target mapping and partitioning for one target do not interfere with the others.

For \textbf{Multi-class Classification} tasks, the pipeline first identifies and filters out rare classes with fewer than two instances to ensure statistical robustness and the feasibility of stratified sampling, subsequently performing an 80/20 train-test split ($random\_state=42$) with stratified sampling to preserve the original class distribution across both subsets.

In the case of \textbf{Regression} tasks, the system incorporates a specialized preprocessing step to handle numerical strings—automatically removing characters such as commas and percentage signs to convert targets into floating-point numbers—before executing a standard 80/20 random split ($random\_state=42$).

For \textbf{Binary Classification tasks}, we employ a consistent 80/20 train-test split utilizing a fixed random seed of 42 to guarantee the reproducibility of our experimental results across different model architectures.

Following the partitioning process, the processed data is structured into standardized DataFrames, with features and target vectors stored separately. These components, along with updated metadata, are persisted to maintain a clear and structured record. This systematic organization ensures absolute input consistency across all models during the subsequent training and evaluation phases.

\subsection{Training Strategy and Early Stopping}

For our NN models, we implement a systematic training protocol that utilizes a validation-based early stopping mechanism to prevent overfitting and ensure optimal generalization.

In contrast to the GBDTs and TabPFN models, which are trained using their respective standard iterations or pre-trained weights without premature termination, all neural models are optimized using a dedicated validation set created by further partitioning the initial training data with a 15\% split ($test\_size=0.15$).

The training process for these neural models utilize a training loop with a maximum of 1,000 epochs and an integrated \texttt{EarlyStopping} callback.

This monitoring strategy tracks the validation loss (\texttt{val\_loss}) with a patience of 30 epochs, meaning the training process is automatically terminated if no improvement in the minimum validation loss is observed for 30 consecutive iterations.

To support this pipeline, numerical and categorical features are processed separately to determine layer dimensions and embedding cardinalities, then converted into tensor format for seamless integration with our model trainers.

\section{Additional Results}

\subsection{Example of a full LLM prompt}
\label{sec:full_prompt}
In the initial screening, we provide dataset and column descriptions as context to an LLM, then ask it to extract the information we need. An example of a full prompt is given below:

% \begin{tcolorbox}[colback=gray!5!white,colframe=gray!75!black]
\begin{mylisting}
Your task is to extract structured information from two text sections: the "About Dataset" section and the file + column descriptions of a Kaggle  dataset page. You will output the information in strict JSON format without adding any explanation or commentary. First, please carefully read the following "About Dataset" section:

<AboutDataset>
"The United States Geological Survey (USGS) dataset provides comprehensive earthquake event data essential for analyzing soil-structure interaction under seismic loading. The dataset includes the following features:

Seismic Event ID: A unique and precise identifier assigned to each earthquake event, ensuring the reliability and accuracy of the data. 
Date and Time: The exact date and time when the seismic event occurred, recorded in the format YYYY-MM-DD HH: MM. 
Magnitude: The magnitude of the earthquake, measured on the Richter scale, indicates the energy released by the seismic event. 
Epicenter Location: The geographical coordinates (latitude and longitude) of the earthquake's epicenter, representing the point on the Earth's surface directly above the earthquake's origin. 
Depth: The depth at which the earthquake originated beneath the Earth's surface, measured in kilometers. 
Peak Ground Acceleration (PGA): The maximum ground acceleration recorded at the site during the earthquake, measured in gravity. 

This dataset is not just a collection of past events, but a critical foundation for the future. It paves the way for predictive modeling and analysis of soil-structure interaction during seismic events. By integrating this data with additional site-specific soil and structural information, researchers can develop machine learning 
models, such as XGBoost, optimized using Particle Swarm Optimization (PSO), to predict and analyze the behavior of soil-structure systems under seismic loading conditions. This enhanced understanding can significantly contribute to the design and safety of structures in earthquake-prone regions, ushering in a new era of seismic safety."
</AboutDataset>

Next, please carefully read the following file + column descriptions:
<FileColumnDescriptions>
  "file_columns": {"top_rated_movies.csv": [
				"adult",
				"backdrop_path",
				"genre_ids",
				"id",
				"original_language",
 				"original_title",
				"overview",
				"popularity",
				"poster_path",
				"release_date",
				"title",
				"video",
				"vote_average",
				"vote_count"
			]}
</FileColumnDescriptions>
When extracting the structured information, please follow these rules 
for each key in the JSON:
- "field": If the method can be clearly determined as one of "CV", "NLP", "ML", select that value. Otherwise, use "OTHERS".
- "task_type": If the task type can be clearly classified into "classification", "regression", "time_series", "recommendation", choose the corresponding type(s). If there are multiple targets, include all relevant types. If the task type cannot be clearly determined, use ["uncertain"].
- "has_multiple_prediction_targets": Set to true if there are multiple
    prediction targets, false otherwise.
- "target_column_names": Select target column name(s) only from the 
    column names provided in "file_columns". If the column names are ex-
    plicitly mentioned as a prediction target, include them. If no co-
    lumn names are explicitly labeled as target, set it to ["uncertain"].
- "derived_from_other_kaggle_dataset": Set to true if it is derived 
    from another Kaggle dataset, false otherwise.
- "appears_on_other_data_websites": List the websites where the data-
    set appears on other data websites. If there are none, use an 
    empty list [].
- "industry": Select one of the following domains: "People and Society", 
    "Business and Finance", "Health and Biology", "Industry", "Nature 
    Science", "Energy", "Transportation", "Human resources", "Internet".

Please output the structured information in strict JSON format immediately without a preamble.
\end{mylisting}
  
\subsection{Metadata Structure}
Our metadata stores key information about the task, including (but not limited to) name, url, number of rows and columns, metafeatures of the corresponding dataset, and evaluation results. The metadata is initially generated from the webpage of dataset, and then we append LLM-processed metadata afterwards, which may cause redundant information. However, we make sure that a few common keys we need will present in this metadata. Furthermore, the metadata is dynamically updated after each model evaluation to include the latest performance scores. 
% , including idx, row, col, categorical\_ratio, mean\_kurtosis, mean\_skewness, mean\_abs\_skewness, target\_entropy, target\_kurtosis, target\_skewness, missing\_rate, industry, and the scores of each models.
An example is given below:
\begin{mylisting}
{
    "idx": "K1057064",
    "about_dataset": "Context
    This dataset contains the responses of a gas multisensor device deployed on the field in an Italian city. Hourly responses averages are recorded along with gas concentrations references from a certified analyzer. This dataset was taken from UCI Machine Learning Repository: https://archive.ics.uci.edu/ml/index.php
    Content
    The dataset contains 9357 instances of hourly averaged responses from an array of 5 metal oxide chemical sensors embedded in an Air Quality Chemical Multisensor Device. The device was located on the field in a significantly polluted area, at road level,within an Italian city. Data were recorded from March 2004 to February 2005 (one year) representing the longest freely available recordings of on field deployed air quality chemical sensor devices responses. Ground Truth hourly averaged concentrations for CO, Non Metanic Hydrocarbons, Benzene, Total Nitrogen Oxides (NOx) and Nitrogen Dioxide (NO2) and were provided by a co-located reference certified analyzer. Evidences of cross-sensitivities as well as both concept and sensor drifts are present as described in De Vito et al., Sens. And Act. B, Vol. 129,2,2008 (citation required) eventually affecting sensors concentration estimation capabilities. Missing values are tagged with -200 value. This dataset can be used exclusively for research purposes. Commercial purposes are fully excluded.
    Attribute Information
    0 Date (DD/MM/YYYY) 
    1 Time (HH.MM.SS) 
    2 True hourly averaged concentration CO in mg/m^3 (reference analyzer) 
    3 PT08.S1 (tin oxide) hourly averaged sensor response (nominally CO targeted) 
    4 True hourly averaged overall Non Metanic HydroCarbons concentration in microg/m^3 (reference analyzer) 
    5 True hourly averaged Benzene concentration in microg/m^3 (reference analyzer) 
    6 PT08.S2 (titania) hourly averaged sensor response (nominally NMHC targeted) 
    7 True hourly averaged NOx concentration in ppb (reference analyzer) 
    8 PT08.S3 (tungsten oxide) hourly averaged sensor response (nominally NOx targeted) 
    9 True hourly averaged NO2 concentration in microg/m^3 (reference analyzer) 
    10 PT08.S4 (tungsten oxide) hourly averaged sensor response (nominally NO2 targeted) 
    11 PT08.S5 (indium oxide) hourly averaged sensor response (nominally O3 targeted) 
    12 Temperature
    13 Relative Humidity (%) 
    14 AH Absolute Humidity",
    "tags": [
        "Earth and Nature",
        "Health",
        "Regression",
        "Environment",
        "Weather and Climate",
        "Pollution"
    ],
    "usability": "10.00",
    "license": "Data files Original Authors",
    "updated_frequence": "Never",
    "url": "https://www.kaggle.com/datasets/fedesoriano/air-quality-data-set",
    "files": [
        "AirQuality.csv"
    ],
    "vote_count": "147",
    "view_count": "161K",
    "file_columns": {
        "AirQuality.csv": [
            "Date;Time;CO(GT);PT08.S1(CO);NMHC(GT);C6H6(GT);PT08.S2(NMHC);NOx(GT);PT08.S3(NOx);NO2(GT);PT08.S4(NO2);PT08.S5(O3);T;RH;AH;;"
        ]
    },
    "method": "ML",
    "task_type": "regression",
    "has_multiple_prediction_targets": true,
    "target_column_names": [
        "CO(GT)",
        "NMHC(GT)",
        "C6H6(GT)",
        "NOx(GT)",
        "NO2(GT)"
    ],
    "derived_from_other_kaggle_dataset": false,
    "appears_on_other_data_websites": [
        "UCI Machine Learning Repository"
    ],
    "domain": "Nature Science",
    "target_columns_trans": [
        "COGT",
        "NMHCGT",
        "C6H6GT",
        "NOxGT",
        "NO2GT"
    ],
    "num_rows": 9357,
    "num_cols": 15,
    "total_missing_rate": 0.0,
    "column_missing_rate": {
        "Date": 0.0,
        "Time": 0.0,
        "COGT": 0.0,
        "PT08S1CO": 0.0,
        "NMHCGT": 0.0,
        "C6H6GT": 0.0,
        "PT08S2NMHC": 0.0,
        "NOxGT": 0.0,
        "PT08S3NOx": 0.0,
        "NO2GT": 0.0,
        "PT08S4NO2": 0.0,
        "PT08S5O3": 0.0,
        "T": 0.0,
        "RH": 0.0,
        "AH": 0.0
    },
    "categorical_columns": [
        "Date",
        "Time",
        "COGT",
        "C6H6GT",
        "T",
        "RH",
        "AH"
    ],
    "numerical_columns": [
        "PT08S1CO",
        "NMHCGT",
        "PT08S2NMHC",
        "NOxGT",
        "PT08S3NOx",
        "NO2GT",
        "PT08S4NO2",
        "PT08S5O3"
    ],
    "target": "NOxGT",
    "hash_tag": "2ef20f50c66d1ad9ed3abfa7acd245cce2db8c1460d074f2ea57938118f27d57",
    "is_duplicated": false,
    "version": "v1.2",
    "industry": "Earth & Environment",
    "categorical_ratio": 0.4667,
    "kurtosis": {
        "PT08S1CO": -0.6369258481978166,
        "NMHCGT": 16.374086051749106,
        "PT08S2NMHC": -0.8186955625501944,
        "NOxGT": -0.30686389356146604,
        "PT08S3NOx": -0.6089245660242302,
        "PT08S4NO2": -0.6628007684432391,
        "PT08S5O3": -0.6997644426785992
    },
    "mean_kurtosis": 1.805730138613366,
    "skewness": {
        "PT08S1CO": 0.42332768950149163,
        "NMHCGT": 4.242236316616238,
        "PT08S2NMHC": 0.22878328916039956,
        "NOxGT": 0.03770772581307186,
        "PT08S3NOx": 0.23275746942300765,
        "NO2GT": -1.3035532964552168,
        "PT08S4NO2": -0.19325248286700444,
        "PT08S5O3": 0.30791186668386217
    },
    "mean_abs_skewness": 0.8711912670650366,
    "target_skew": -1.3035532964552168,
    "target_kurt": 0.18248575061633376,
    "Field": "ML",
    "lgb": 0.7543906120776662,
    "CatBoost": 0.7818531684331507,
    "XgBoost": 0.79929119348526,
    "FTTransformer": 0.7167628908206897,
    "resnet": 0.5734872817993164,
    "mlp": 0.06815792347112881,
    "realmlp": 0.5921510404339285,
    "tabpfn": 0.7132938504219055
}
\end{mylisting}

\subsection{Fingerprints}
\label{sec:fingerprint}
To ensure the integrity of our experimental results and prevent redundant evaluations, we implement a systematic deduplication protocol that generates a unique digital fingerprint for every dataset in our benchmark.

The fingerprinting process initiates by constructing a structural descriptor string composed of the dataset's dimensions ($rows \times columns$), a sorted list of all feature column names to ensure permutation invariance, and specific statistical characteristics of the target variable tailored to the identified task type.

This comprehensive metadata string is processed using the SHA-256 cryptographic hashing algorithm to produce a fixed-length hexadecimal digest, serving as a robust and unique identifier for each specific data configuration.

By comparing these generated hashes across the entire repository, we are able to identify and exclude functionally identical datasets even if they are named differently or stored in different locations, thereby maintaining a high-quality, non-redundant experimental corpus.

Following the verification of uniqueness, each dataset is registered within our pipeline, ensuring that every downstream evaluation is conducted on a distinct set of features and labels. We give two examples of fingerprints as follows: 

An example of classification dataset, where we use class counts to quantify the target column distribution:
\begin{mylisting}
470x17 | cols:AGE,DGN,PRE10,PRE11,PRE14,PRE17,PRE19,PRE25,PRE30,PRE32,PRE4,PRE5,PRE6,PRE7,PRE8,PRE9,Risk1Yr | target:70,400
\end{mylisting}
which is hashed by \textit{hashlib.sha256}
\begin{mylisting}
66f0fcd8e73457529c8956ac2970fb29b99f234887982d0e31b9bfebde3cce65
\end{mylisting}

An exmaple of regression dataset, where we use deciles along with min and max values to quantify the target column distribution:
\begin{mylisting}
2000x61 | cols:V1,V10,V11,V12,V13,V14,V15,V16,V17,V18,V19,V2,V20,V21,V22,V23,V24,V25,V26,V27,V28,V29,V3,V30,V31,V32,V33,V34,V35,V36,V37,V38,V39,V4,V40,V41,V42,V43,V44,V45,V46,V47,V48,V49,V5,V50,V51,V52,V53,V54,V55,V56,V57,V58,V59,V6,V60,V7,V8,V9,class | target:0.0,36.0,69.8,104.0,139.0,175.0,211.0,247.3,284.0,320.0,354.0
\end{mylisting}
which is hased by \textit{hashlib.sha256}
\begin{mylisting}
1bcd7c11429b234d05711e1915064a68f52fe6801af58a1aba853b955a366d46
\end{mylisting}

\subsection{Full pairwise comparison of model performance gap}
\label{sec:more_gaps}
\begin{figure*}[!ht]
    \centering
    \includegraphics[width=0.98\linewidth]{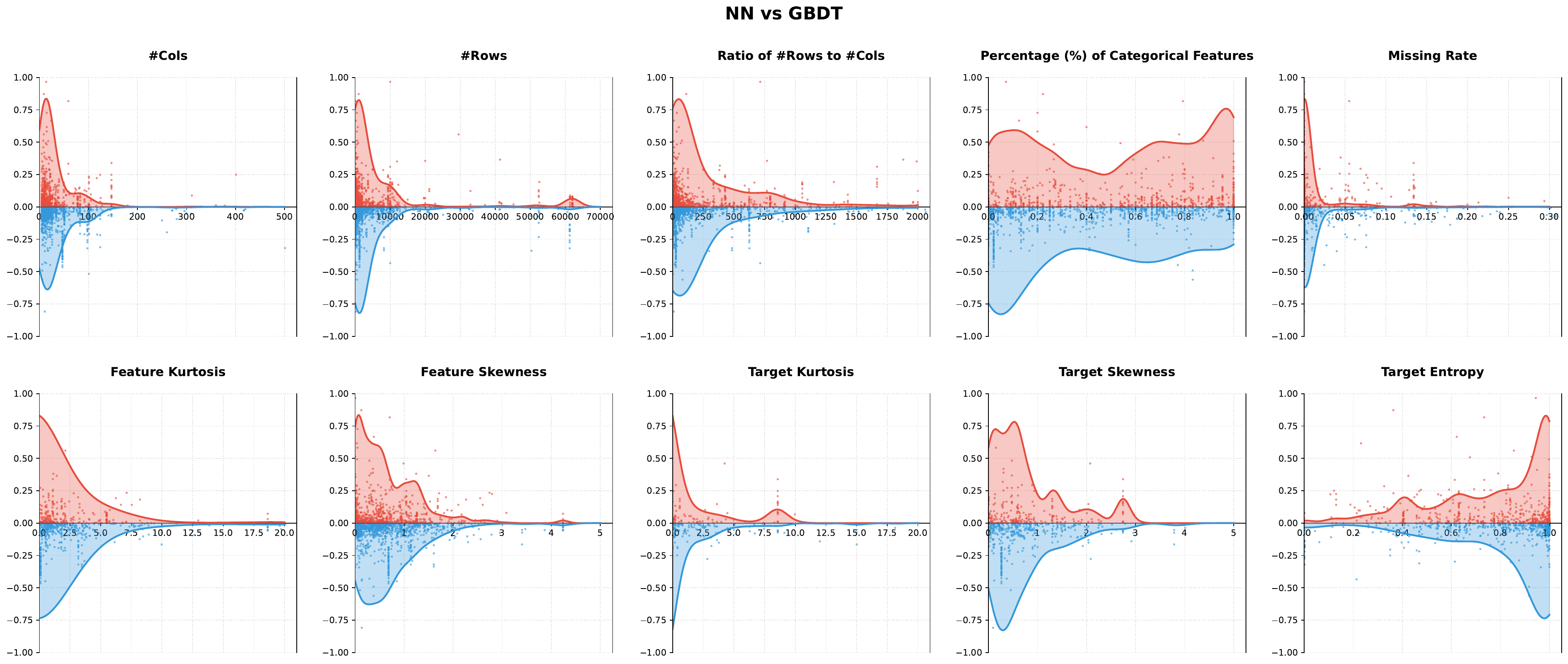}
    \caption{The distribution of performance gap between NNs and GBDT. Performance gap on different datasets refers to the subtraction of the score between the former and latter models (for example, NN vs GBDT means subtracting score of GBDT from NN), which are quantified by red (positive/win) and blue (negative/loss) points, respectively. We also fit a PDF of the points along each varying metafeatures via kernel density estimation.}
    \label{fig:NN_vs_GBDT}
\end{figure*}

\begin{figure*}[!ht]
    \centering
    \includegraphics[width=0.98\linewidth]{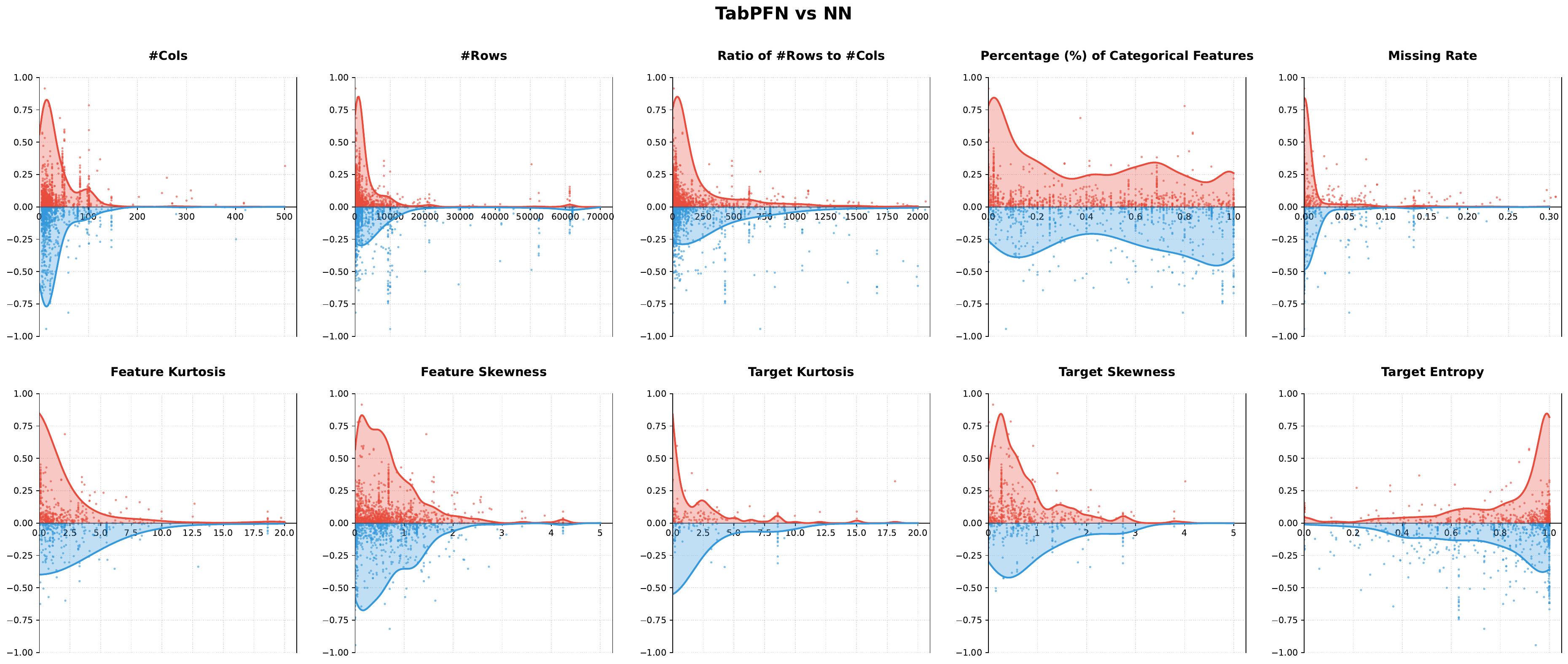}
    \caption{The distribution of performance gap between TabPFN and NN. }
    \label{fig:TabPFN_vs_NN}
\end{figure*}

\begin{figure*}[!ht]
    \centering
    \includegraphics[width=0.98\linewidth]{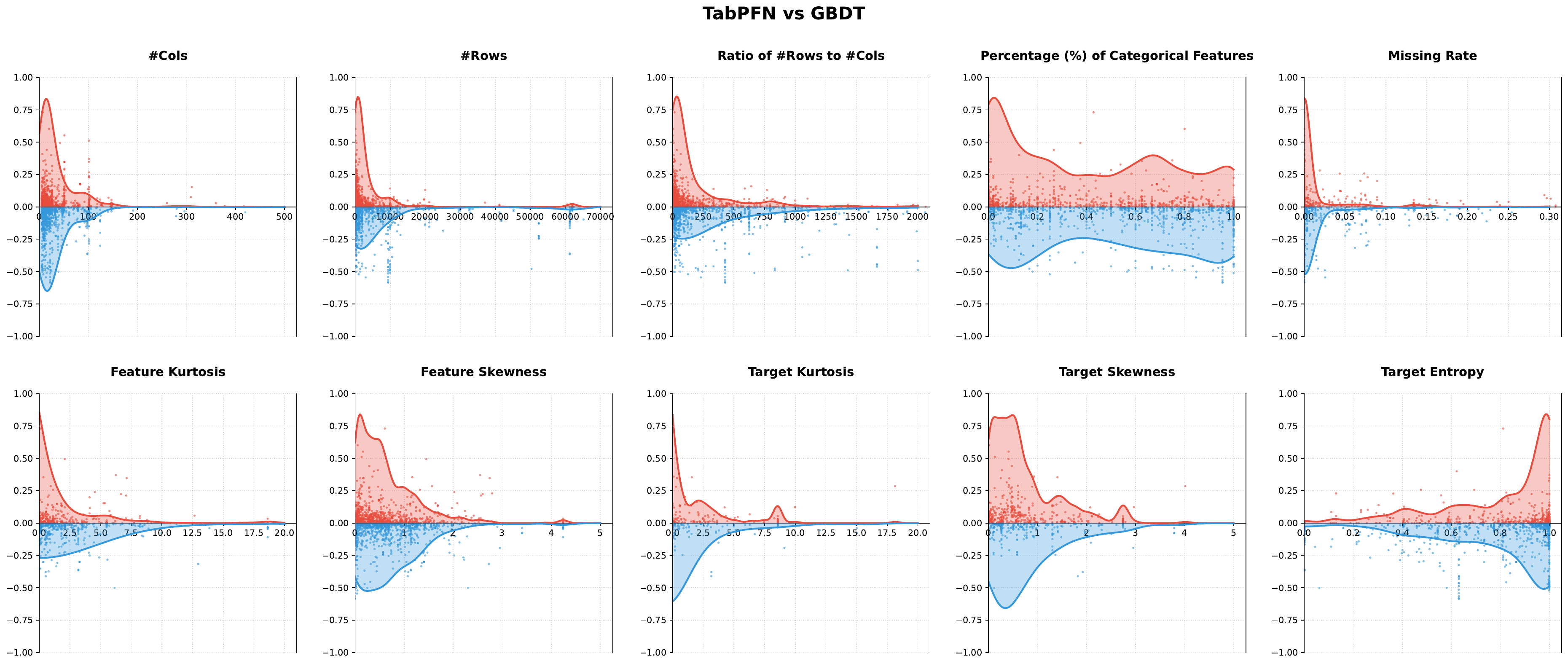}
    \caption{The distribution of performance gap between TabPFN and GBDT. }
    \label{fig:TabPFN_vs_GBDT}
\end{figure*}

For completeness, \Cref{fig:NN_vs_GBDT,fig:TabPFN_vs_NN,fig:TabPFN_vs_GBDT} visualize the pairwise model performance gap along all metafeatures.

\end{document}